\documentclass{article}
\usepackage{spconf}
\usepackage{xcolor,amsmath,graphicx,booktabs,arydshln}
\usepackage{hyperref}
\usepackage{makecell}
\usepackage[numbers]{natbib}

\usepackage{multirow    }

\newcommand\slimeq{\mkern1.5mu{=}\mkern1.5mu}

\title{Contrastive prediction strategies for unsupervised segmentation\\and categorization of phonemes and words}
\name{%
\begin{tabular}{c}%
Santiago Cuervo$^1$,\qquad Maciej Grabias$^1$,\qquad Jan Chorowski$^{2}$,\qquad Grzegorz Ciesielski$^1$ \\ Adrian {\L}a{\'n}cucki$^3$,\qquad Paweł Rychlikowski$^1$,\qquad Ricard Marxer$^4$%
\end{tabular}%
\thanks{%
The authors thank the Polish National Science Center for funding
under the OPUS-18 2019/35/B/ST6/04379 grant and the PlGrid consortium for computational resources. We also thank the French National Research Agency for their support through the ANR-20-CE23-0012-01 (MIM) grant.}%
}%
\address{$^1$ University of Wroclaw, Poland \quad
$^2$NavAlgo, France\quad $^3$NVIDIA, Poland\\
$^4$ Universit\'e de Toulon, Aix Marseille Univ, CNRS, LIS, Toulon, France}
\copyrightnotice{\begin{minipage}{\textwidth}\scriptsize \copyright \quad 2021 IEEE. Personal use of this material is permitted. Permission from IEEE must be obtained for all other uses, in any current or future media, including  reprinting/republishing this material for advertising or promotional purposes, creating new collective works, for resale or redistribution to servers or lists, or reuse of any copyrighted component of this work in other works.\end{minipage}}

\begin{document}

\maketitle

\begin{abstract}
We identify a performance trade-off between the tasks of phoneme categorization and phoneme and word segmentation in several self-supervised learning algorithms based on Contrastive Predictive Coding (CPC). 
Our experiments suggest that context building networks, albeit necessary for high performance on categorization tasks, harm segmentation performance by causing a temporal shift on the learned representations. Aiming to tackle this trade-off,
we take inspiration from the leading approaches on segmentation and propose multi-level Aligned CPC (mACPC).
It builds on Aligned CPC (ACPC), a variant of CPC which exhibits the best performance on categorization tasks,
and incorporates multi-level modeling and optimization for detection of spectral changes.
Our methods improve in all tested categorization metrics and achieve state-of-the-art performance in word segmentation.

\end{abstract}
\section{Introduction}
\label{sec:intro}
Speech self-supervised learning (SSL) without linguistic labels is aimed at producing representations that are useful for downstream problems such as transcription, classification or understanding. The prior work focuses mainly on either automatic boundary detection of phonemes or words \cite{kreuk2020selfsupervised, bhati2021segmental, kamper21a}, or learning representations which expose phonemic information~\cite{Oord18a, chorowski21a}, and facilitate phoneme prediction with linear transformations. Even though the tasks seem related, an approach which performs well on the former may do poorly on the latter. An extreme example is an encoding that alternates between only two labels at every phone change. This representation has the full information about the boundaries, yet no information about the sequence of phonemes. On the other hand, we may imagine an encoding with no abrupt changes at phoneme boundaries, in which every frame maps to the correct phoneme through some unknown linear projection.


Contrastive Predicting Coding (CPC)~\cite{Oord18a} and its variants are popular methods of approaching these tasks. CPC is an SSL algorithm which extracts latent representations from sequential data by learning to predict future states of the model. An encoder $g_{enc}$ maps consecutive overlapping chunks of data to latent representations $z$, producing sequences of codes. An autoregressive model $g_{ar}$ is then applied to the latent representations and trained to predict $M$ upcoming latents. The model is trained using Noise-Contrastive Estimation (NCE): the prediction $p_t$ of the latent code $z_t$ at time $t$ must be closer to $z_t$ than to randomly sampled latent codes, termed negative samples. When applied to speech, CPC produces acoustic representations which are useful for phoneme prediction and low-resource speech recognition.

Different variations of CPC have been proposed in the literature and have shown improvements on various downstream tasks. \citet{chorowski21a} presented Aligned CPC (ACPC), in which rather than producing individual predictions for each future representation, the model emits a sequence of $K < M$ predictions which are aligned to the $M$ upcoming representations. In this way, $g_{ar}$ solves a simpler task of predicting the next symbols, but not their exact timing, while $g_{enc}$ is incentivized to produce piece-wise constant latent codes. ACPC exhibits higher linear phone prediction accuracy and lower ABX error rates than CPC, while being slightly faster to train due to the reduced number of prediction heads.


CPC-based techniques have also been applied to unsupervised phoneme and word segmentation. \citet{kreuk2020selfsupervised} proposed a model that omits the prediction network $g_{ar}$ and is instead trained to discriminate between adjacent and non-adjacent representations. The model learns to detect spectral changes in the signal and tends to produce piece-wise constant latent codes. Boundaries are obtained as peaks 
in cosine dissimilarity between consecutive latent representations. 

Segmental CPC (SCPC)~\cite{bhati2021segmental} improves upon \cite{kreuk2020selfsupervised} by adding 
a standard CPC feature extractor with $M\slimeq 1$ to model the signal at the level of segments of frames. A differentiable implementation of the boundary detector used by \cite{kreuk2020selfsupervised} is applied to the frame-level latent representations and those within boundaries are averaged to produce a sequence of segment representations that is fed as input to another feature extractor. The segment-level feature extractor is meant to operate roughly at phoneme level and act as a language model. At test time the boundary detector from \cite{kreuk2020selfsupervised} is used at the frame-level to predict phoneme boundaries and at the segment level for word boundaries. SCPC reported state-of-the-art performance in both phone and word unsupervised segmentation.


We investigate the performance of ACPC on phoneme segmentation, and of \cite{kreuk2020selfsupervised} and SCPC on phoneme classification accuracy and ABX. The results suggest a trade-off between segmentation and categorization performance. We then explore the variations to the standard CPC model to understand the causes of this conflict. We also propose a multi-level ACPC model aiming to obtain gains in segmentation performance similar to SCPC  \cite{kreuk2020selfsupervised}, and explore the effect of multi-level modelling on phoneme classification and the ABX task. Finally, we show that including an auxiliary contrastive loss between adjacent frames as in \cite{kreuk2020selfsupervised} in ACPC models consistently improves segmentation and categorization performance.

\begin{figure}[hb]
    \centering
    \includegraphics[width=0.49\textwidth]{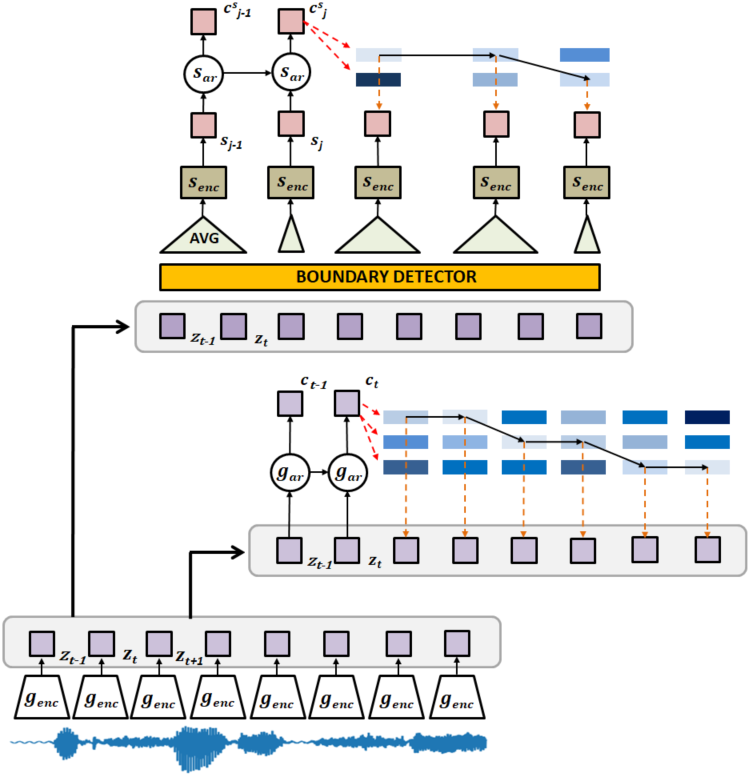}
    \caption{The mACPC model has two main modules: frame-level and segment level. The frame-level module works on raw waveforms and extracts latent representations. These are processed by the boundary detector, which predicts boundaries and averages latents within those boundaries to produce segment representations. Finally, the segment-level module learns to predict higher-level features.}
    \vspace{-10pt}
    \label{fig:hacpc}
\end{figure}

\section{Augmenting ACPC for segmentation}

We propose multi-level ACPC (mACPC; Fig. \ref{fig:hacpc}),
which extends the ACPC architecture with a second ACPC feature extractor to model the signal at the level of frame segments, similarly to \cite{bhati2021segmental}.
At the frame level, a strided convolutional encoder $g_{enc}$ maps the input sequence $x_1, \dots, x_T$ to a sequence of encoded frames $z_1, \dots, z_{T'}$. The encoded frames are used to determine the position of the segment boundaries. Following \cite{kreuk2020selfsupervised}, we calculate the score for placing a boundary at the position $t$ as $-\text{sim}(z_{t-1}, z_{t})$, where $\text{sim}(\cdot)$ denotes cosine similarity. Peaks in the dissimilarity scores indicate boundaries. Frames within two consecutive boundaries are considered as segments and their constituents are averaged. We obtain the final segment representations $s_1, \dots, s_{J}$ by feeding the averages to a segment-level encoder $s_{enc}$. While the frame-level auto-regressive model $g_{ar}$ summarizes all encoded frames up to time $t$ into a context vector $c_t = g_{ar}(z_{\leq t})$, the auto-regressive model $s_{ar}$ does it at the segment level $c^{s_j} = s_{ar}(s_{\leq j})$. Finally, $K$ and $K^s$ predictions are made at frame and segment levels conditioned on the corresponding context vectors, which are then aligned to $M$ and $M^s$ upcoming encoded frames and segments respectively. The ACPC prediction loss, as described in \cite{chorowski21a}, is applied at both levels. The two prediction losses from frames and segments are summed into the total loss to be optimized.

Additionally, we also consider variations of ACPC and mACPC in which we add to their total loss the contrastive loss between adjacent representations proposed by Kreuk et al. in \cite{kreuk2020selfsupervised} to optimize for detection of spectral changes in the signal. The loss is applied to the output of $g_{enc}$.







\section{Experiments}
\label{sec:experiments}



We perform two phone classification evaluations: frame-wise classification, and an alignment-insensitive evaluation using Connectionist Temporal Classification (CTC) \cite{ctc}. The classifiers are trained on the LibriSpeech train-clean-100 dataset \cite{ls100} for 10 epochs. For the frame-wise case, a linear classifier is optimized with a cross-entropy loss and we report accuracy. The model used for the CTC evaluation is a single-layer bidirectional LSTM network with 256 hidden units, followed by a 1D convolution with 256 filters, kernel width 8 and stride of 4, trained using the CTC loss in which emission of the blank character is forbidden to force classifying each frame as a phoneme. Performance is evaluated using Phoneme Error Rate (PER).


Phoneme segmentation experiments are run on both TIMIT \cite{timit} and Buckeye \cite{PITT200589}. Word segmentation is only run on Buckeye as in \cite{bhati2021segmental}. Segmentation quality was measured with precision, recall, F1-score and over-segmentation robust R-value \cite{rvalue}, with a 20ms tolerance \cite{kreuk2020selfsupervised, bhati2021segmental}.

We follow the methodology from \cite{kreuk2020selfsupervised, bhati2021segmental} for the train/test split and pre-processing of the corpora, and train our models on the union of LibriSpeech \emph{train-clean-100} and the \emph{train} split of Buckeye. All models are trained for 50 epochs using a minibatch size of 32.  




\begin{figure}[t!]
    \centering
    \includegraphics[width=0.46\textwidth]{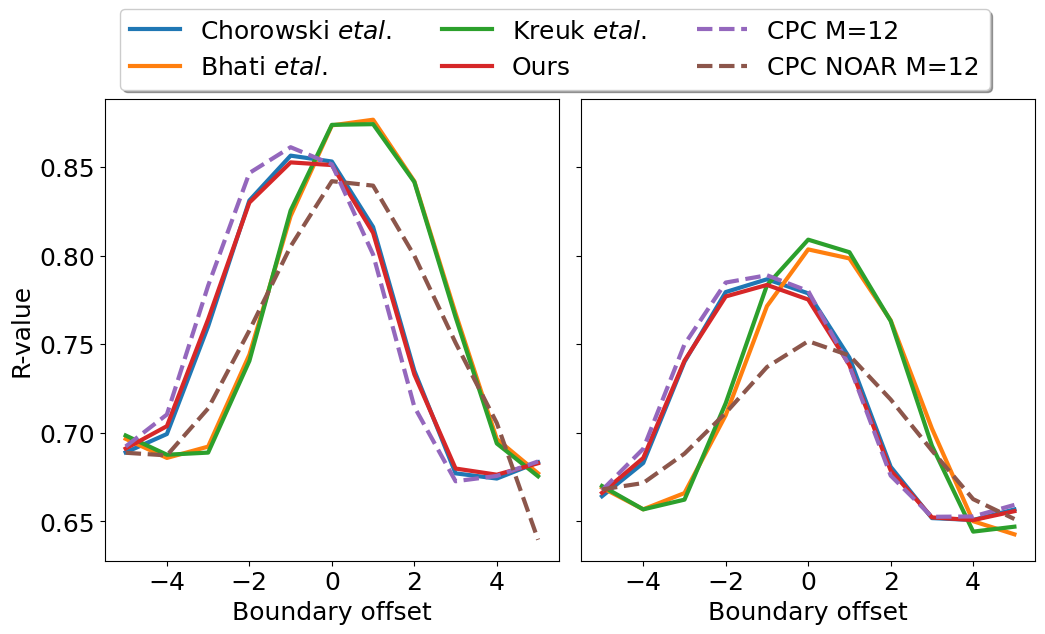}
    \caption{Segmentation performance for different predicted boundary shifts on TIMIT (left) and Buckeye (right). Models with context builders perform better with an offset, indicating a representation shift.}
    \label{fig:offset_rvalues}
\end{figure}

All models read single channel raw waveforms sampled at 16kHz, chunked into sequences of 20480 samples. Encoder $g_{enc}$ applies five 1D convolutions with internal dimension 256 and filter widths (10; 8; 4; 4; 4). Convolutions are followed by channel-wise magnitude normalization and all but the last by ReLU activations. Convolutions are strided by (5; 4; 2; 2; 2), resulting in a 160-fold rate reduction, yielding a 256-dimensional latent vector extracted every 10ms. The segment is a network with a single fully connected hidden layer of 512 units and ReLU activation. Context-building models $g_{ar}$ and $s_{ar}$ are two-layer LSTM networks \cite{lstm} with 256 units. We use $K\slimeq 6$, $M\slimeq 12$, $K^{s}\slimeq 2$ and $M^{s}\slimeq 4$ for (m)ACPC models. Each prediction head accesses all past contexts through a single transformer layer \cite{transformer} with 8 scaled dot-product attention heads with internal dimension 2048 and dropout \cite{dropout} with $p = 0.1$. 

Our implementation is available at \url{https://github.com/chorowski-lab/CPC_audio}.

\sthanks{}


\section{Results and discussion}
\label{sec:results}

\subsection{Shifted predictions due to context building} 
Upon visual inspection of CPC and ACPC segmentations, we observed that predicted boundaries are often shifted in the same direction and by a similar amount. We hypothesize that the use of a context building network in the CPC model promotes pushing into the future the representation of the underlying signal. This would ease prediction of several steps ahead especially when the recurrent layer of the context builder is capable of exploiting the increased past context.

To test this possibility we re-evaluate the segmentation performance of the different methods by offsetting the predicted boundary positions by several fixed values. Note that each offset evaluates on different amount of data due to border effects, however scores are averaged across whole utterances limiting the consequences of this difference. Figure~\ref{fig:offset_rvalues} shows how ACPC and mACPC exhibit an optimal offset value at 1 frame step (-10ms), while other methods peak at 0ms as expected from predictions without misalignment issues. To ensure this effect is due to the context builder and not other changes in ACPC (or mACPC), we also run a CPC model with and without it (dotted lines). It is worth noting that the optimal offset for both of these techniques is the same for the two datasets, however further investigation is needed to know whether this value is independent from the data. This consistent offset can be considered as a hyperparameter of the model. For comparison purposes, in the following experiments we report segmentation scores at a fixed offset of -10ms for models with a context builder.

\begin{table}[t]
\setlength{\tabcolsep}{3pt}
\caption{Phone segmentation on TIMIT (top) and Buckeye (bottom) test sets}
\centering
\def\arraystretch{0.5}
\begin{tabular}{lcccc}
\toprule
Model & Precision & Recall & F1 & R-val \\ \midrule
Kreuk et al. & 84.80 & \textbf{85.77} & 85.27 & 87.35 \\
SCPC & \textbf{85.31} & 85.36 & \textbf{85.31} & \textbf{87.38} \\
ACPC & 83.41 & 83.15 & 83.26 & 85.64 \\
ACPC + Kreuk et. al. loss & 83.68 & 84.74 & 84.69 & 86.86 \\
mACPC & 82.53 & 83.05 & 82.78 & 85.26 \\
mACPC  + Kreuk et. al. loss & 84.63 & 84.79 & 84.70 & 86.86 \\
\midrule
Kreuk et al. & 76.27 & 78.42 & 77.31 & 80.35 \\
SCPC & \textbf{77.21} & \textbf{78.95} & \textbf{78.03} & \textbf{80.90} \\
ACPC & 74.44 & 76.28 & 75.32 & 78.66 \\
ACPC + Kreuk et. al. loss & 74.68 & 76.59 & 75.59 & 78.88 \\
mACPC & 74.00 & 76.04 & 74.98 & 78.34 \\
mACPC + Kreuk et. al. loss & 74.70 & 76.81 & 75.72 & 78.97 \\ \bottomrule
\end{tabular}
\label{tab:seg}
\label{tab:phone_seg}
\vspace{-1.5mm}
\caption{Phone classification on the test split of LibriSpeech train-clean-100}
\centering
\setlength{\tabcolsep}{4pt}
\def\arraystretch{0.5}
\begin{tabular}{@{}lcccc@{}}
\toprule
 & \multicolumn{2}{c}{On $z$ vectors} & \multicolumn{2}{c}{On $c$ vectors} \\
 \cmidrule(lr){2-3} \cmidrule(lr){4-5}
Model & Acc. & PER & Acc. & PER \\ \midrule
Kreuk et al. & 44.87 & 32.46 & - & - \\
SCPC & 43.79 & 31.62 & - & - \\
ACPC & 47.62 & 24.34 & 67.87 & 18.10 \\
\makecell[l]{ACPC + Kreuk et. al. loss} & 47.82 & 25.93 & 67.99 & 18.15 \\
mACPC  & 50.98 & \textbf{21.15} & 69.97 & 16.91 \\ 
\makecell[l]{mACPC + Kreuk et. al. loss} & \textbf{51.64} & 21.69 & \textbf{70.25} & \textbf{16.65} \\ \bottomrule
\end{tabular}
\label{tab:phone_class}
\end{table}

\subsection{Comparative study on segmentation and classification of phonemes}


Tables~\ref{tab:phone_seg} and \ref{tab:phone_class} show the performance of the studied methods on phoneme segmentation and  classification, respectively. Segmentation performance increases by adding to the model in \cite{kreuk2020selfsupervised} a second CPC at the segment level (as in \cite{bhati2021segmental}). Interestingly ACPC \cite{chorowski21a} and mACPC do not attain the same segmentation performance level despite their similarities and the offset correction. On the other hand they do achieve much better phoneme prediction rates, both frame synced (frame-wise accuracy) and through alignment (CTC PER). This points to an apparent compromise between ensuring proper boundaries and retaining phonemic information.

\begin{table}[t!]
\centering
\caption{Classification and phone segmentation performance of CPC variants. Frame-wise accuracy is calculated on encoded frames.}
\setlength{\tabcolsep}{2pt}
\def\arraystretch{0.5}
\begin{tabular}{@{}lccc@{}}
\toprule
 &  & \multicolumn{2}{c}{R-value} \\
\cmidrule{3-4}
Model & Frame Acc. & Buckeye & TIMIT \\
\midrule
CPC, $M\slimeq 1$ & 17.28 & 66.98 & 70.77 \\
CPC, $M\slimeq 12$, no $g_{ar}$ & 38.11 & 75.17 & 84.20 \\ 
CPC, $M\slimeq 12$ & 47.90 & 78.89 & 86.11 \\
\bottomrule
\end{tabular}
\label{tab:compromise}
\end{table}


In order to assess which factors influence this trade-off, we perform a study in which each of the identified changes to CPC are tested\footnote{We do not report variations for the number of negative samples because they did not have a significant effect on any of the metrics} (Table~\ref{tab:compromise}). 
The one step ahead prediction used by \cite{kreuk2020selfsupervised} and SCPC is not in itself enough to improve segmentation performance. The key aspect, as hinted in \cite{kreuk2020selfsupervised}, is the removal of the frame level context builder and the use of the encoded frames themselves as predictors.

Furthermore, the models in which we augment the ACPC loss with an auxiliary contrastive loss between adjacent representations show consistent improvements on segmentation and classification metrics, motivating its use as a method to tackle the performance trade-off. Moreover, the representations obtained by these models did not present a time shift even when using a context builder. This indicates that optimizing for detection of spectral changes penalizes representation shifting, and at least partially explains the better segmenting ability of \cite{kreuk2020selfsupervised} and SCPC.

\subsection{The ZeroSpeech ABX task}
Phoneme segmentation and classification are often targeted due to their potential use in 
applications such as the ABX task from the ZeroSpeech challenge \cite{zs2021}, in which the objective is to match a speech example to its equivalent by choosing from two others that differ in a single phone. In this situation conserving the phonemic content is crucial since it is the discriminant factor. Results in Table~\ref{tab:abx} confirm that methods such as mACPC that excel in frame-wise accuracy and CTC PER also do so in ABX.

\subsection{Integration of higher-level structure}

We observe that the addition of higher level information using the second head in mACPC improves the results. SCPC showed an improvement in the case of phoneme segmentation, here we show the benefit of such strategy in phoneme discrimination with mACPC. Furthermore, it enables the detection of word boundaries using the technique from \cite{bhati2021segmental}. Word boundary experiment results in Table~\ref{tab:word_seg} indicate that mACPC outperforms SCPC. We hypothesize that the superior phoneme representations of mACPC improve the quality of the pseudo-language model.  

We also analyse topline performance in phoneme categorization and word-segmentation of SCPC and mACPC by using ground-truth phoneme boundaries during training. For mACPC, oracle segmentations improve PER on encoded frames from $21.15$ (Table \ref{tab:phone_class}) to $20.12$ and word segmentation R-value from $47.40$ (Table \ref{tab:word_seg}) to $49.20$. This contrasts with SCPC where use of ground-truth segments marginally affects results ($31.62$ to $31.93$ PER and $45.39$ to $46.1$ R-value), suggesting that the representations obtained by the contrastive loss on adjacent frames  \cite{kreuk2020selfsupervised} are insufficient for language modelling. The improved performance of mACPC when using ground-truth segmentation should motivate further work on improving estimated segmentation.

\begin{table}[!t]
\centering
\caption{ABX scores on ZeroSpeech 2021 dev set. For models using context-networks the values are calculated on context vectors.}
\def\arraystretch{0.5}
\begin{tabular}{@{}lcc@{}}
\toprule

Model & \multicolumn{1}{c}{ABX within} & \multicolumn{1}{c}{ABX across} \\ \midrule
Kreuk et al. & 10.93 & 19.11 \\
SCPC & 20.18 & 16.26 \\
ACPC & 5.78 & 7.93 \\
ACPC + Kreuk et. al. loss & 5.67 & 7.78 \\
mACPC & 5.28 & 7.13 \\ 
mACPC + Kreuk et. al. loss & \textbf{5.13} & \textbf{6.84} \\\bottomrule
\end{tabular}
\label{tab:abx}
\vspace{-0.5mm}
\caption{Word segmentation on Buckeye test set}
\centering
\setlength{\tabcolsep}{3pt}
\def\arraystretch{0.5}
\begin{tabular}{@{}lcccc@{}}
\toprule
Model & Precision & Recall & F1 & R-val \\ \midrule
SCPC & 36.23 & \textbf{32.75} & 34.33 & 45.39 \\
mACPC & \textbf{42.06} & 30.32 & \textbf{35.05} & \textbf{47.40} \\
mACPC + Kreuk et. al. loss & 40.36 & 30.86 & 34.83 & 47.11 \\\bottomrule
\end{tabular}
\label{tab:word_seg}
\end{table}

\section{Conclusions and future work}
\label{sec:illust}
We investigate the applicability of CPC-based models to unsupervised phoneme segmentation and classification. As in \cite{bhati2021segmental} we propose a two-level model acting simultaneously on time synchronous frames and on variable-length segments, to capture linguistic regularities. We discover a performance compromise between phoneme segmentation and classification, and find that it stems from a roughly constant prediction shift induced by CPC's context modeling. We show that this issue can be alleviated by manually removing the offset from the representations or by using an auxiliary contrastive loss between consecutive latent representations. After accounting for the prediction shift, our model achieves competitive segmentation performance and outperforms existing approaches in phoneme classification, transcription, word boundary detection and ABX tests. Furthermore, the use of oracle phonemic alignments indicate that improvements on segment estimation may lead to even better performance in these tasks.




\bibliographystyle{IEEEtranN}
\bibliography{strings,refs}

\begin{thebibliography}{14}
\providecommand{\natexlab}[1]{#1}
\providecommand{\url}[1]{#1}
\csname url@samestyle\endcsname
\providecommand{\newblock}{\relax}
\providecommand{\bibinfo}[2]{#2}
\providecommand{\BIBentrySTDinterwordspacing}{\spaceskip=0pt\relax}
\providecommand{\BIBentryALTinterwordstretchfactor}{4}
\providecommand{\BIBentryALTinterwordspacing}{\spaceskip=\fontdimen2\font plus
\BIBentryALTinterwordstretchfactor\fontdimen3\font minus
  \fontdimen4\font\relax}
\providecommand{\BIBforeignlanguage}[2]{{%
\expandafter\ifx\csname l@#1\endcsname\relax
\typeout{** WARNING: IEEEtranN.bst: No hyphenation pattern has been}%
\typeout{** loaded for the language `#1'. Using the pattern for}%
\typeout{** the default language instead.}%
\else
\language=\csname l@#1\endcsname
\fi
#2}}
\providecommand{\BIBdecl}{\relax}
\BIBdecl

\bibitem[Kreuk et~al.(2020)Kreuk, Keshet, and Adi]{kreuk2020selfsupervised}
F.~Kreuk, J.~Keshet, and Y.~Adi, ``{Self-Supervised Contrastive Learning for
  Unsupervised Phoneme Segmentation},'' in \emph{Proc. Interspeech 2020}, 2020,
  pp. 3700--3704.

\bibitem[Bhati et~al.(2021)Bhati, Villalba, Żelasko, Moro-Velázquez, and
  Dehak]{bhati2021segmental}
S.~Bhati, J.~Villalba, P.~Żelasko, L.~Moro-Velázquez, and N.~Dehak,
  ``{Segmental Contrastive Predictive Coding for Unsupervised Word
  Segmentation},'' in \emph{Proc. Interspeech 2021}, 2021, pp. 366--370.

\bibitem[Kamper and van Niekerk(2020)]{kamper21a}
\BIBentryALTinterwordspacing
H.~Kamper and B.~van Niekerk, ``Towards unsupervised phone and word
  segmentation using self-supervised vector-quantized neural networks,''
  \emph{CoRR}, vol. abs/2012.07551, 2020. [Online]. Available:
  \url{https://arxiv.org/abs/2012.07551}
\BIBentrySTDinterwordspacing

\bibitem[van~den Oord et~al.(2018)van~den Oord, Li, and Vinyals]{Oord18a}
\BIBentryALTinterwordspacing
A.~van~den Oord, Y.~Li, and O.~Vinyals, ``Representation learning with
  contrastive predictive coding,'' \emph{CoRR}, vol. abs/1807.03748, 2018.
  [Online]. Available: \url{http://arxiv.org/abs/1807.03748}
\BIBentrySTDinterwordspacing

\bibitem[Chorowski et~al.(2021)Chorowski, Ciesielski, Dzikowski, Łańcucki,
  Marxer, Opala, Pusz, Rychlikowski, and Stypułkowski]{chorowski21a}
J.~Chorowski, G.~Ciesielski, J.~Dzikowski, A.~Łańcucki, R.~Marxer, M.~Opala,
  P.~Pusz, P.~Rychlikowski, and M.~Stypułkowski, ``{Aligned Contrastive
  Predictive Coding},'' in \emph{Proc. Interspeech 2021}, 2021, pp. 976--980.

\bibitem[Graves et~al.(2006)Graves, Fern\'{a}ndez, Gomez, and Schmidhuber]{ctc}
\BIBentryALTinterwordspacing
A.~Graves, S.~Fern\'{a}ndez, F.~Gomez, and J.~Schmidhuber, ``Connectionist
  temporal classification: Labelling unsegmented sequence data with recurrent
  neural networks,'' in \emph{Proceedings of the 23rd International Conference
  on Machine Learning}, ser. ICML '06.\hskip 1em plus 0.5em minus 0.4em\relax
  New York, NY, USA: Association for Computing Machinery, 2006, p. 369–376.
  [Online]. Available: \url{https://doi.org/10.1145/1143844.1143891}
\BIBentrySTDinterwordspacing

\bibitem[Panayotov et~al.(2015)Panayotov, Chen, Povey, and Khudanpur]{ls100}
V.~Panayotov, G.~Chen, D.~Povey, and S.~Khudanpur, ``Librispeech: An asr corpus
  based on public domain audio books,'' in \emph{2015 IEEE International
  Conference on Acoustics, Speech and Signal Processing (ICASSP)}, 2015, pp.
  5206--5210.

\bibitem[Garofolo et~al.(1993)Garofolo, Lamel, Fisher, Fiscus, Pallett, and
  Dahlgren]{timit}
J.~S. Garofolo, L.~F. Lamel, W.~M. Fisher, J.~G. Fiscus, D.~S. Pallett, and
  N.~L. Dahlgren, ``Darpa timit acoustic phonetic continuous speech corpus
  cdrom,'' 1993.

\bibitem[Pitt et~al.(2005)Pitt, Johnson, Hume, Kiesling, and
  Raymond]{PITT200589}
M.~A. Pitt, K.~Johnson, E.~Hume, S.~Kiesling, and W.~Raymond, ``The buckeye
  corpus of conversational speech: labeling conventions and a test of
  transcriber reliability,'' \emph{Speech Communication}, vol.~45, no.~1, pp.
  89--95, 2005.

\bibitem[Räsänen et~al.(2009)Räsänen, Laine, and Altosaar]{rvalue}
O.~J. Räsänen, U.~K. Laine, and T.~Altosaar, ``{An improved speech
  segmentation quality measure: the r-value},'' in \emph{Proc. Interspeech
  2009}, 2009, pp. 1851--1854.

\bibitem[Hochreiter and Schmidhuber(1997)]{lstm}
S.~Hochreiter and J.~Schmidhuber, ``{Long Short-Term Memory},'' \emph{Neural
  Computation}, vol.~9, no.~8, pp. 1735--1780, 11 1997.

\bibitem[Vaswani et~al.(2017)Vaswani, Shazeer, Parmar, Uszkoreit, Jones, Gomez,
  Kaiser, and Polosukhin]{transformer}
A.~Vaswani, N.~Shazeer, N.~Parmar, J.~Uszkoreit, L.~Jones, A.~N. Gomez, L.~u.
  Kaiser, and I.~Polosukhin, ``Attention is all you need,'' in \emph{Advances
  in Neural Information Processing Systems}, I.~Guyon, U.~V. Luxburg,
  S.~Bengio, H.~Wallach, R.~Fergus, S.~Vishwanathan, and R.~Garnett, Eds.,
  vol.~30.\hskip 1em plus 0.5em minus 0.4em\relax Curran Associates, Inc.,
  2017.

\bibitem[Srivastava et~al.(2014)Srivastava, Hinton, Krizhevsky, Sutskever, and
  Salakhutdinov]{dropout}
N.~Srivastava, G.~Hinton, A.~Krizhevsky, I.~Sutskever, and R.~Salakhutdinov,
  ``Dropout: A simple way to prevent neural networks from overfitting,''
  \emph{Journal of Machine Learning Research}, vol.~15, no.~56, pp. 1929--1958,
  2014.

\bibitem[Nguyen et~al.(2020)Nguyen, de~Seyssel, Rozé, Rivière, Kharitonov,
  Baevski, Dunbar, and Dupoux]{zs2021}
T.~A. Nguyen, M.~de~Seyssel, P.~Rozé, M.~Rivière, E.~Kharitonov, A.~Baevski,
  E.~Dunbar, and E.~Dupoux, ``The zero resource speech benchmark 2021: Metrics
  and baselines for unsupervised spoken language modeling,'' 2020.

\end{thebibliography}

\end{document}